\documentclass{article}

\pdfoutput=1
\usepackage{PRIMEarxiv}

\usepackage[utf8]{inputenc} % allow utf-8 input
\usepackage[T1]{fontenc}    % use 8-bit T1 fonts
\usepackage{hyperref}       % hyperlinks
\usepackage{url}            % simple URL typesetting
\usepackage{booktabs}       % professional-quality tables
\usepackage{amsfonts}       % blackboard math symbols
\usepackage{nicefrac}       % compact symbols for 1/2, etc.
\usepackage{microtype}      % microtypography
\usepackage{lipsum}
\usepackage{fancyhdr}       % header
\usepackage{graphicx}       % graphics
\graphicspath{{media/}}     % organize your images and other figures under media/ folder

\usepackage{times}  % DO NOT CHANGE THIS
\usepackage{helvet}  % DO NOT CHANGE THIS
\usepackage{courier}  % DO NOT CHANGE THIS

\usepackage{graphicx} % DO NOT CHANGE THIS
\usepackage{caption} % DO NOT CHANGE THIS AND DO NOT ADD ANY OPTIONS TO IT
\usepackage{subfigure}
\usepackage{amsmath,amssymb,amsfonts}
\usepackage{algorithm}
\usepackage{algorithmic}
\usepackage{color}
\usepackage{soul}
\usepackage{newfloat}
\usepackage{listings}
\title{A template for Arxiv Style
\thanks{\textit{\underline{Citation}}: 
\textbf{Authors. Title. Pages.... DOI:000000/11111.}} 
}

\author{
  Yachao Yang,Yanfeng Sun, Shaofan Wang,  Fujiao Ju, Baocai Yin \\
  Beijing Key Laboratory of Multimedia and Intelligent Software Technology\\
  Beijing Institute of Artificial Intelligence\\
  Faculty of Information Technology, Beijing University of Technology\\
  Beijing, China \\
  \texttt{\{yangyc\}@emails.bjut.edu.cn} \\
    \texttt{\{yfsun, wangshaofan, jfj2017, ybc\}@bjut.edu.cn} \\
   \And
 Jipeng Guo\\
 College of Information Science and Technology\\
 Beijing University of Chemical Technology\\
 Beijing 100029, China.
  \texttt{\{guojipeng\}@bupt.edu.cn} \\
     \And
 Junbin Gao\\
Discipline of Business Analytics\\
The University of Sydney Business School\\
The University of Sydney, Camperdown, NSW 2006, Australia\\
  \texttt{\{junbin.gao\}@sydney.edu.au} \\
}

\begin{document}
\maketitle

\begin{abstract}
Graph Neural Networks (GNNs) have excelled in handling graph-structured data, attracting significant research interest. However, two primary challenges have emerged: interference between topology and attributes distorting node representations, and the low-pass filtering nature of most GNNs leading to the oversight of valuable high-frequency information in graph signals. These issues are particularly pronounced in heterophilic graphs. To address these challenges, we propose Dual-Frequency Filtering Self-aware Graph Neural Networks (DFGNN). DFGNN integrates low-pass and high-pass filters to extract smooth and detailed topological features, using frequency-specific constraints to minimize noise and redundancy in the respective frequency bands. The model dynamically adjusts filtering ratios to accommodate both homophilic and heterophilic graphs. Furthermore, DFGNN mitigates interference by aligning topological and attribute representations through dynamic correspondences between their respective frequency bands, enhancing overall model performance and expressiveness. Extensive experiments conducted on benchmark datasets demonstrate that DFGNN outperforms state-of-the-art methods in classification performance, highlighting its effectiveness in handling both homophilic and heterophilic graphs.
\end{abstract}

\section{Introduction}

Graphs data are widely present in various domains, including social, chemical and financial networks. GNNs are powerful tools for extracting valuable information from complex graph data by aggregating neighborhood information of objective node and capturing inherent linkages and patterns within the data.  Moreover, GNNs are crucial in many applications relying on structural dependencies, such as node classification~\cite{qiu2024refining30}, link prediction~\cite{li2024evaluating11}, and recommender systems~\cite{wang2024unleashing12}.

GNNs can be categorized into spatial domain-based and spectral domain-based types. Spatial domain-based GNNs aggregate neighbour representations based on the spatial structure of nodes, acting as low-pass filters to smooth representations. However, for heterophilic graphs, which are common in the real world and often have missing or spurious edges~\cite{wang2021graph4}, capturing high-frequency information that highlights node differences may be more effective~\cite{bo2021beyond13}. This paper focuses on spectral domain-based GNNs, which filter graph signals through spectral filters in the Fourier domain, theoretically allowing for the extraction of any frequency information. Methods like Cheby-GCN~\cite{defferrard2016convolutional2}, BernNet~\cite{he2021bernnet14}, and GPRGNN~\cite{chien2020adaptive15} use high-order polynomials to approximate arbitrary filters, which are better at processing heterophilic graphs but come with high computational complexity. Additionally, GNN-L/HF~\cite{zhu2021interpreting16} filters attributes in different frequencies but remains a low-pass filter for topological embedding, lacking adaptability to different graph structures. Although FAGCN~\cite{bo2021beyond13} and PC-Conv~\cite{li2024pc34} mix high-frequency and low-frequency feature representations, they blur the distinction between these frequencies in the model.

Meanwhile, most spectral-based methods aim to achieve a unified representation that fits attribute representations while satisfying structural constraints~\cite{ma2021unified6}. However, recent studies have shown that interference between structure and attributes affects this unified representation. Inaccuracies in structural information further exacerbate this issue in heterophilic graphs. To address this problem, GNNBC~\cite{yang2022graph7} employs two separate representations to independently learn attribute and topological information. However, GNNBC overlooks the interdependence between attributes and structure within the same data, leading to minimal performance improvement on homophilic graphs. In contrast, GNN-SATA~\cite{yang8} uses soft association constraints to explore the relationship between features and structure, achieving strong performance on both homophilic and heterophilic graphs. Nevertheless, the model has insufficient constraints on attributes, and the hyperparameter settings for different datasets are complex.

To effectively extract diverse frequency information and alleviate interference between typology and attributes, we propose a new model called Dual-Frequency Filtering Self-aware Graph Neural Networks (DFGNN) for both homophilic and heterophilic graphs. DFGNN enhances representation learning by integrating low-pass and high-pass filters at the topological level, capturing both smooth representations and detailed features. Sparse and low-rank constraints are applied to reduce noise and redundant information in their respective frequency bands. DFGNN dynamically adjusts high-pass and low-pass filtering ratios to adapt to different graph structures automatically. Furthermore, DFGNN establishes a frequency-specific information interaction between filtered topological and attribute representations, aligning them to mitigate interference effectively. This alignment enhances the model's expressiveness and flexibility, leveraging relevant frequency information from both domains. %This mechanism aligns the topological representation with the attribute representation, enhancing the model's expressiveness and flexibility by effectively utilizing relevant frequency information from both domains. 
Consequently, the iterative updating way between attribute and structural representations further enhances DFGNN's comprehensive modelling capability for graph data. The main contributions of this paper are summarized as follows:
\begin{itemize}
\item From an optimization perspective, DFGNN employs self-aware low-pass and high-pass filters to capture smooth and detailed features, allowing the model to adaptively learn specific representations suitable for homophilic and heterophilic graphs. Furthermore, frequency-specific constraints are applied to enhance the capture of structural information.
\item DFGNN achieves effective alignment and interaction between attribute and topological representations by establishing dynamic associations between their corresponding frequency bands.
\item Extensive experiments conducted on both homophilic and heterophilic benchmark datasets demonstrate that DFGNN outperforms state-of-the-art methods in classification performance.
\end{itemize}
% \begin{figure*}[t!]
% \begin{center}
%  {\includegraphics[width=176mm,height=50mm]{framework.pdf}}
% \end{center}
% \caption{An overview of the proposed GNN-SATA model. }\label{framework}
% \end{figure*} 
\section{Relate Works}
\subsection{Spectral Graph Neural Networks}
Spectral graph neural networks leverage spectral graph theory to process graph-structured data by performing graph convolutions using the eigenvalues and eigenvectors of the graph Laplacian matrix. Due to the high computational complexity of eigenvalue decomposition, Defferrard et al. \cite{defferrard2016convolutional2} proposed ChebNet to reduce computational complexity. Subsequently, Kipf et al. \cite{kipf2017semi3} introduced the Graph Convolutional Network (GCN), which further simplifies computations using a first-order approximation of ChebNet. As a result, GCN has become one of the most widely used GNN architectures. Recent studies ~\cite{ma2021unified6,zhu2021interpreting16} indicate that the optimization objective of GCN can be expressed as follows: 
\begin{equation}
\min _{\mathbf{Z}}\operatorname{tr} \left(\mathbf{Z}^{\top} \tilde{\mathbf{L}} \mathbf{Z}\right) ,\mathbf{Z} = \mathbf{X}\mathbf{W}^*
    \label{os}
\end{equation}
where $\widetilde{\mathbf{L}}$ is the normalized graph Laplacian matrix, $\mathbf{X}$ is the feature matrix, $\mathbf{W}^* = \mathbf{W}^{0}\cdots\mathbf{W}^{L-1}$ are weight matrix, $L$ is the layer number of GCN, and $\operatorname{tr}(\cdot)$ represents the matrix trace operation. When $L$ is too large, the well-known over-smoothing problem arises. Researchers have proposed several methods based on residual or jump connections to address this, such as APPNP~\cite{gasteiger2018predict32}, GCNII~\cite{chen2020simple29}, and JKNet~\cite{xu2018representation33}. These methods establish connections between encoded representations and attributes, alleviating information loss caused by multi-layer propagation. The optimization equations for these methods can be expressed as follows:
       \begin{equation}
\min _{\mathbf{Z}} \left\| \mathbf{Z}-\mathbf{X}\right\|_F^2+\alpha_1 \operatorname{tr}\left(\mathbf{Z}^{\top} \tilde{\mathbf{L}} \mathbf{Z}\right)
          \label{oa}
      \end{equation}
      
Recent studies indicate that using a single $\mathbf{Z}$ to learn both attribute and topological representations can cause interference between structure and attributes. To address this, two new networks, GNNBC and GNN-SATA, have been proposed with the following optimization objectives:
       \begin{equation}
\min _{\mathbf{F,Z}} \left\| \mathbf{F}-\mathbf{X}\right\|_F^2+\alpha_1 \operatorname{tr}\left(\mathbf{Z}^{\top} \tilde{\mathbf{L}} \mathbf{Z}\right)+\alpha_2 G(\mathbf{F},\mathbf{Z},\mathbf{A})
          \label{ob}
      \end{equation}
They use $\mathbf{F}$ to learn attribute representations and $\mathbf{Z}$ for structural representations, where $G(\cdot)$ denotes the relationship between $\mathbf{F}$, $\mathbf{Z}$ and $\mathbf{A}$. GNNBC uses the Hilbert-Schmidt independence criterion (HSIC)~\cite{gretton2005measuring9} loss to disentangle $\mathbf{F}$ and $\mathbf{Z}$. However, in homophilic graphs, where attribute and structural representations are inherently interconnected, the HSIC constraint may lead to the loss of crucial information, resulting in moderate performance. In contrast,  GNN-SATA dynamically connects the attribute representation $\mathbf{F}$ with the adjacency relationship $\mathbf{A}$, enabling the adjacency relationship to be adaptively learned. %making the adjacency relationship dynamically learnable. 
GNN-SATA demonstrates promising results on both homophilic and heterophilic datasets. However, GNN-SATA primarily focuses on exploring structural relationships and overlooks feature constraints, while its hyperparameter tuning for different datasets remain complex.  % are complicated.

\subsection{Homophilic and Heterophilic Graphs}
Homophilic graphs are characterized by nodes with similar features or labels that are more likely to be connected. In contrast, heterophilic graphs have nodes with different features or labels that are more likely to be connected. Traditional GNNs typically operate under the homophily assumption, which limits their effectiveness for heterophilic graphs. Recent methods have been developed based on structural exploration~\cite{jin2020graph5,he2022block17,liu2023beyond18,pei2020geom24,qiu2024refining19}, adversarial approaches~\cite{zhu2019robust20,zhang2020gnnguard21,zhang2020self22,suresh2021adversarial23}, and frequency information extraction~\cite{bo2021beyond13,zhu2021interpreting16} to address these challenges. 

Structure exploration methods focus on discovering sparse, low-rank, or high-order comprehensive graph relationships, ultimately forming a more accurate and robust graph structure. Adversarial methods enhance the model's robustness against malicious perturbations by simulating adversarial attacks, thereby improving the model's security and stability. Frequency information extraction methods use different filters to capture information from various frequency bands, optimizing the processing of both homophilic and heterophilic graph data. However, these frequency-based methods typically consider only attribute or structural information when handling graphs. This incomplete separation of information limits the model's comprehensive understanding of complex graph structures.

\begin{figure*}[!t]
\begin{center}
 {\includegraphics[width=150mm,height=70mm]{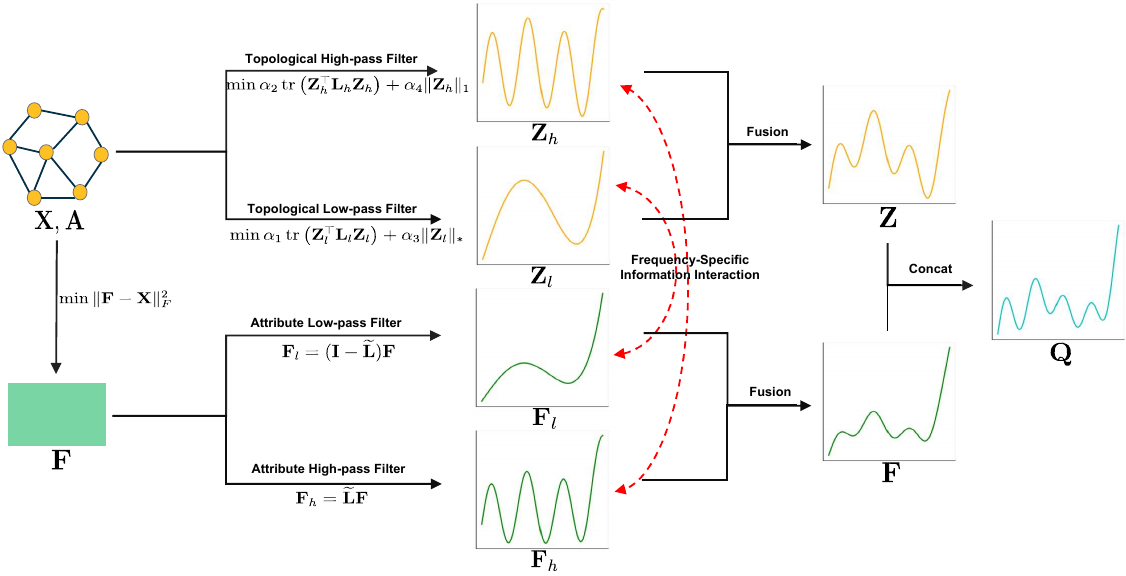}}
\end{center}
\caption{An overview of the proposed DFGNN model. }\label{framework}
\end{figure*}

\section{Model Formulation}\label{Sec:4}
\subsection{Notations}
For an attribute graph $\mathcal{G}=\{\mathcal{V},\mathcal{E}\}$, $\mathcal{V}$ and $\mathcal{E}$ represent the sets of $N$ nodes and $E$ edges, respectively. The node features are denoted by $\mathbf{X}\in \mathbb{R}^{N\times d}$, and the adjacency relationships are captured by the adjacency matrix $\mathbf{A}\in \mathbb R^{N\times N}$. When there is an edge between nodes $v_i$ and $v_j$, $\mathbf{A}_{ij}=1$; otherwise, $\mathbf{A}_{ij}=0$. The normalized graph Laplacian matrix mentioned above is defined as $\widetilde{\mathbf{L}}=\mathbf{I}-\widetilde{\mathbf{A}}$, where $\widetilde{\mathbf{A}}=\bar{\mathbf{D}}^{-\frac{1}{2}}\bar{\mathbf{A}}\bar{\mathbf{D}}^{-\frac{1}{2}}$ is the normalized adjacency matrix,  $\bar{\mathbf{A}}= \mathbf{A}+\mathbf{I}$ is an adjacency matrix that includes self-connections,  $\bar{\mathbf{D}}=\mathbf{D}+\mathbf{I}$, where $\mathbf{D}$ is the degree diagonal matrix of $\mathbf{A}$ with $\mathbf{D}_{ii} = \sum_{j} \mathbf{A}_{ij}$. For semi-supervised classification tasks, only a subset of nodes $\mathcal{V}_L$ has the corresponding label set $\mathcal Y_L$. We aim to predict the labels of the unlabeled nodes subset $\mathcal{V}_N$, i.e., $\mathcal{V} = \mathcal{V}_L \cup \mathcal{V}_N$.

\subsection{The Proposed DFGNN}
Low-pass filters enhance the features of similar nodes by smoothing the information of neighbouring nodes. This 
improves feature aggregation and produces smoother feature representations of graph data. In contrast, high-pass filters highlight differences between connected nodes by emphasizing the high-frequency components of the graph signal, allowing the model to distinguish nodes with connected edges but different classes. This distinction is crucial for preserving the unique attributes of each node and capturing the structural details of heterophilic graphs.

\textbf{Step1: Capturing dual-frequency graph information.} Real-world graph structures are rarely purely homophilic or heterophilic; they usually contain a mix of both types of connections. This makes it insufficient to use only low-pass or high-pass filters. The model can extract complementary information by applying topological low-pass and high-pass filters. Based on this motivation and Eq.\eqref{os}, we first propose the following optimization formula:
           \begin{equation}
      	\begin{split}
&\min _{\mathbf{Z}_l,\mathbf{Z}_h} \alpha_1 \operatorname{tr}\left(\mathbf{Z}_l^{\top} {\mathbf{L}_{l}}\mathbf{Z}_l\right)+ \alpha_2 \operatorname{tr}\left(\mathbf{Z}_h^{\top} {\mathbf{L}_{h}}\mathbf{Z}_h\right)
           	\end{split}
            \label{losseq1}
      \end{equation}
where $\alpha_1, \alpha_2$  are adaptive adjustment parameters for different datasets. $\mathbf{L}_{l} = \widetilde{\mathbf{L}}$ functions as a low-pass graph filter. Minimizing $ \operatorname{tr}\left(\mathbf{Z}_l^{\top} {\mathbf{L}_{l}}\mathbf{Z}_l\right)$  emphasizes low-frequency information by smoothing the signal and enhancing the similarities between connected nodes, which are captured in the low-frequency components. In contrast, $ \mathbf{L}_{h} = \mathbf{I} - \widetilde{\mathbf{L}}$  acts as a high-pass graph filter. Minimizing $ \operatorname{tr}\left(\mathbf{Z}_h^{\top} {\mathbf{L}_{h}}\mathbf{Z}_h\right)$  highlights high-frequency information by reducing the similarities between connected nodes and accentuating their differences, which are captured in the high-frequency components.

\textbf{Step2: Regularizing dual-frequency graph information.} Low-frequency information captures the global structure of the graph and common features among connected nodes. Low-rank constraints on the feature matrix make low-frequency information smoother and more representative, thereby improving the model's ability to capture node similarities, which is essential for dealing with homophilic graphs. In contrast,  high-frequency information primarily captures the graph's differences between nodes and local structures. Sparse constraints reduce redundant features, highlight critical local information, and enhance the model's ability to distinguish node differences. This approach is essential for handling heterophilic graphs. By different regularization to constraint low-frequency and high-frequency information, we can better capture both local and global structural information, thereby enhancing the representation capability of the graph neural network. Thus, the regularized model can be expressed as follows:
           \begin{equation}
      	\begin{split}
&\min _{\mathbf{Z}_l,\mathbf{Z}_h}\alpha_1 \operatorname{tr}\left(\mathbf{Z}_l^{\top} {\mathbf{L}_{l}}\mathbf{Z}_l\right)+ \alpha_2 \operatorname{tr}\left(\mathbf{Z}_h^{\top} {\mathbf{L}_{h}}\mathbf{Z}_h\right)
          \\
    &+\alpha_3\|\mathbf{Z}_l\|_*+\alpha_4\|\mathbf{Z}_h\|_1.
           	\end{split}
            \label{loss23}
      \end{equation}
where $\|\cdot\|_*$ and $\|\cdot\|_1$ represent the $l_1$ norm and nuclear norm, respectively, and $\alpha_3$ and $\alpha_4$ are also the corresponding adaptive balance parameters.

\textbf{Step3: Linking attribute and topological representations.} As the number of network layers increases, Eq.~\eqref{loss23} will encounter over-smoothing problems. Inspired by Eq.~\eqref{ob}, DFGNN introduces a separate attribute representation $\mathbf{F}$, which helps mitigate the over-smoothing issue and reduce the interference between attributes and topology. Different from previous frequency domain decomposition methods that only consider attribute or topological, the attribute representation is decomposed into high-frequency and low-frequency components through attribute filters, aligning the frequencies of attribute and topological representations for joint optimization. The proposed DFGNN is formulated as follows:
           \begin{equation}
      	\begin{split}
&\min _{\mathbf F, \mathbf{Z}_l,\mathbf{Z}_h}\left\| \mathbf{F}-\mathbf{X}\right\|_F^2+ \alpha_1 (\operatorname{tr}\left(\mathbf{Z}_l^{\top} {\mathbf{L}_{l}}\mathbf{Z}_l\right)+\left\| \mathbf{F}_{l}-\mathbf{Z}_l \right\|_F^2)
\\&+ \alpha_2 (\operatorname{tr}\left(\mathbf{Z}_h^{\top} {\mathbf{L}_{h}}\mathbf{Z}_h\right)+\left\| \mathbf{F}_{h}-\mathbf{Z}_h \right\|_F^2)
       +\alpha_3\|\mathbf{Z}_l\|_*+\alpha_4\|\mathbf{Z}_h\|_1.
           	\end{split}
            \label{loss22}
      \end{equation}
where low-frequency part of the attribute representation is given by $\mathbf{F}_{l}=(\mathbf{I}-\widetilde{\mathbf{L}})\mathbf{F}$, while the high-frequency part is $\mathbf{F}_{h}=\widetilde{\mathbf{L}}\mathbf{F}$.  The parameters \(\alpha_1, \alpha_2, \alpha_3, \alpha_4\) are adaptively adjusted based on the loss of downstream tasks.

\subsection{Loss Function}
DFGNN combines attribute and topology representations for node classification tasks. Specifically, we fuse dual-frequency representations $\mathbf{Z}_l$ and $\mathbf{Z}_h$ using a simple summation: $\mathbf{Z} = \beta_1 \mathbf{Z}_l + \beta_2 \mathbf{Z}_h$, where $\beta_1$ and $\beta_2$ are self-aware parameters for different datasets. Then, we concatenate the attribute representation $\mathbf{F} = \mathbf{F}_{l} + \mathbf{F}_{h}$ with the fused topology representation $\mathbf{Z}$ to form the final representation $\mathbf{Q}= [\mathbf{F} \| \mathbf{Z}]$, where $\|$ denotes the concatenation operation. This concatenated representation is then passed through a linear transformation and a Softmax function for node classification tasks:

\begin{equation}
     \mathbf{Q}^\prime = softmax(\mathbf{Q}\cdot\mathbf {W_\mathbf{Q} +b})
    \label{softmax}
\end{equation}
where $\mathbf{Q}^\prime_{ij}$ represents the probability that the $i$-th node belongs to the $j$-th class. The final cross-entropy loss is given by:
\begin{equation}
{\mathcal{L}}= \sum_{y_i \in\mathcal Y_L} {\ell \;(\mathbf{Q}_i^\prime,\mathbf{y}_i)}
\label{l2}
\end{equation}
where $\mathbf{y}_i$ is the one-hot encoding of the label of the $i$-th node, and $\ell(\cdot)$ represents the cross-entropy loss. The framework of the model is shown in Figure~\ref{framework}.
\begin{algorithm}[tb]
\caption{Optimization Algorithm of DFGNN}
\label{alg:algorithm}
\textbf{Input}: Feature matrix $\mathbf{X}$, adjacency matrix $\mathbf{A}$, maximum iteration $K$, label set $\mathcal Y_L$.  \\
\textbf{Output}: Classification results $\mathbf{Q}$.\\
\textbf{Intalize Parameters}: $\mathbf{W}_\mathbf{Q}, \mathbf{W}_\mathbf{F}$, $\alpha_1$, $\alpha_2$, $\alpha_3$, $\alpha_4$, $\beta_1$, $\beta_2$.

\begin{algorithmic}[1] %[1] enables line numbers
\STATE Initializing $\mathbf{F}^0$ ,$\mathbf{Z}_l^0$ and $\mathbf{Z}_h^0$.
\WHILE{\emph{iter}$<$1000 and tolerate$<$patience}
\FOR{\emph{k} = 0, 1, $\cdots$, $K-1$}
\STATE Calculate\;$\mathbf{F}^{k+1},\mathbf{Z}_l^{k+1}, \mathbf{Z}_h^{k+1}$\;by\;Eq.\eqref{a19},
 \eqref{a20},\eqref{a21},\eqref{311},\eqref{322}.
\ENDFOR
\STATE Calculate $\mathbf{Q}=[\mathbf{F}^{K}\|\mathbf{Z}^{K}]$
\STATE Calculate classification loss by Eq.\eqref{l2}
\STATE Update parameter $\mathbf{W}_\mathbf{Q}, \mathbf{W}_\mathbf{F}$ , $\alpha_1$, $\alpha_2$, $\alpha_3$, $\alpha_4$, $\beta_1$, $\beta_2$ by gradient descent.
\ENDWHILE
\end{algorithmic}
\end{algorithm}
\begin{table}[t]
\centering
\caption{The statistics of eight datasets.}
\scriptsize
\begin{tabular}{llllll}
\hline
Dataset & \#Nodes & \#Edges & \#Features & \#Classes & HR \\ \hline
\texttt{Cora} & 2708 & 5429 & 1433 & 7 & 0.81 \\
\texttt{Citeseer} & 3327 & 4732 & 3703 & 6 & 0.72 \\
\texttt{Photo} & 7650 & 119081 & 745 & 8 & 0.82 \\
\texttt{Computer} & 13752 & 245861 & 767 & 10 & 0.79 \\
\texttt{Squirrel} & 5201 & 217073 & 2089 & 5 & 0.20 \\ 
\texttt{Chameleon} & 2277 & 36101 & 2325 & 5 & 0.23 \\
\texttt{Wisconsin} & 251 & 499 & 1703 & 5 & 0.11 \\
\texttt{Texas} & 183 & 309 & 1703 & 5 & 0.21 \\
\hline
\end{tabular}
\label{dataset1}
\end{table}
\subsection{Model Optimization}
To determine $\mathbf{F}$, $\mathbf{Z}_l$, and  $\mathbf{Z}_h$, we employ alternating direction iteration to optimize the objective function Eq.\eqref{loss22} except for non-differentiable $l_1$-norm and nuclear norm. %we will utilize alternative strategies to address these terms later. 
Therefore, after removing two frequency-specific constraints, the partial derivatives with respect to $\mathbf{F}$, $\mathbf{Z}_l$, and  $\mathbf{Z}_h$ are:
\begin{align}
&\mathbf{F} - \mathbf{X} + \alpha_1\widetilde{\mathbf{A}}^{\top}(\bar{\mathbf{A}}\mathbf{F}- \mathbf{Z}_l) +\alpha_2 \widetilde{\mathbf{L}}^{\top}(\widetilde{\mathbf{L}}\mathbf{F} - \mathbf{Z}_h)  =0
\\
&   \alpha_1 \widetilde{\mathbf{L}} \mathbf{Z}_l + \alpha_1 (\mathbf{Z}_l - \widetilde{\mathbf{A}}\mathbf{F})=0\\
&  \alpha_2 \bar{\mathbf{A}} \mathbf{Z}_h + \alpha_2 (\mathbf{Z}_h - \widetilde{\mathbf{L}}\mathbf{F})=0
\end{align}
where $\widetilde{\mathbf{A}} = \mathbf{I} - \widetilde{\mathbf{L}}$ as mentioned before and the iterative connection between the model's $k$ th and $(k+1)$th layers is represented as:
\begin{align}
&\mathbf{F}^{k+1}=\mathbf X-\alpha_1\widetilde{\mathbf{A}}^{\top}(\widetilde{\mathbf{A}}\mathbf{F}^{k}- \mathbf{Z}_l)-\alpha_2\widetilde{\mathbf{L}}^{\top}(\widetilde{\mathbf{L}}\mathbf{F}^{k} - \mathbf{Z}_h)\label{a16}
\\
& \mathbf{Z}_l^{k+1}=(\widetilde{\mathbf{A}}\mathbf{Z}_l^{k}+\widetilde{\mathbf{A}}\mathbf{F}^{k})/2\label{a17}
\\
& \mathbf{Z}_h^{k+1} = \widetilde{\mathbf{L}}\mathbf{F} -\widetilde{\mathbf{A}}\mathbf{Z}_h^{k}  \label{a18}
\end{align}
then we try to combine the above optimization algorithms and deep neural networks to optimize the training process~\cite{li2018optimization39}. We introduce additional training parameters $\mathbf{W}_\mathbf{F}$, the activation function $\sigma = \text{ReLU}$. The following is the enhanced iteration formula:
\begin{align}
	\begin{split}
&\mathbf{F}^{k+1}=\mathbf X-\sigma[\mathbf{W_F}(\alpha_1\widetilde{\mathbf{A}}^{\top}\widetilde{\mathbf{A}}+\alpha_2\widetilde{\mathbf{L}}^{\top}\widetilde{\mathbf{L}})]\mathbf{F}^{k}+ \alpha_1\widetilde{\mathbf{A}}^{\top}\mathbf{Z}_l
          \\
    &+\alpha_2\widetilde{\mathbf{L}}^{\top}\mathbf{Z}_h
           	\end{split}\label{a19}\\
&\mathbf{Z}_l^{k+1}=(\sigma(\widetilde{\mathbf{A}}\mathbf{Z}_l^{k})+\widetilde{\mathbf{A}}\mathbf{F}^{k})/2\label{a20}\\
&\mathbf{Z}_h^{k+1} = \widetilde{\mathbf{L}}\mathbf{F} -\sigma(\widetilde{\mathbf{A}}\mathbf{Z}_h^{k})\label{a21}
\end{align}    

To minimize the \(l_1\)-norm and nuclear norm constraints, we use the Forward-Backward splitting methods mentioned in~\cite{combettes2011proximal25}. $\mathbf{Z}_h$ and $\mathbf{Z}_l$ can be updated one more step after Eq.~\eqref{a21} as follows:
\begin{align}
 \mathbf{Z}_l^{k+1}&=\operatorname{prox}_{\alpha_3\|\cdot\|_{*}}(\mathbf{Z}_l^{k+1})\label{311}
\\
\mathbf{Z}_h^{k+1}& =\operatorname{prox}_{\alpha_4\|\cdot\|_{1}}(\mathbf{Z}_h^{k+1}) \label{322}
\end{align}
and the proximal operators for the nuclear norm and the \(l_1\)-norm are represented as follows:
% \GaoC{It seems these steps are added arbitrarily. Reviewers may raise concerns.  Is it possible to directly apply alternating scheme for (6)?}
% \GaoC{Here (18) gives the solution to the proximal problem $\min \|\mathbf Z - \mathbf Z_l\|^2_F + \alpha_3 \|\mathbf Z\|_*$, not sure how to explain why this step can be applied. Similarily (19) is for $\min \|\mathbf Z - \mathbf Z_h\|^2_F + \alpha_4 \|\mathbf Z\|_1$. Have you seen any papers doing this way?}\Yang{Response:~\cite{jin2020graph5} uses the same approach to achieve sparse and low-rank constraints}\GaoC{Yes, you may cite Jin's paper here. Also the idea originates from Hugo Raguet, Jalal Fadili, and Gabriel Peyré. 2013. A generalized forward-
% backward splitting. SIAM Journal on Imaging Sciences 6, 3 (2013), 1199–1226.  Better to clearly show how you get to this way.  They call this the Incremental Proximal
% Descent method.  }
\begin{align}
&\operatorname{prox}_{\alpha_3\|\cdot\|_*}(\mathbf{Z}_l)=\mathbf{U} \operatorname{diag}\left(\left(\gamma_i-\alpha_3\right)_{+}\right) \mathbf{V}^T\label{nuc} \\
&\operatorname{prox}_{\alpha_4\| \cdot\|_1}(\mathbf{Z}_h)=\operatorname{sgn}(\mathbf{Z}_h) \odot(|\mathbf{Z}_h|-\alpha_4)_{+}\label{l1}
\end{align} 
where $\mathbf{U}\operatorname{diag}\left(\gamma_1, \cdots,\gamma_N\right) \mathbf{V}^\top$ denote the singular value decomposition of $\mathbf{Z}_l$. Algorithm~\ref{alg:algorithm} summarizes the entire optimization process of DFGNN.
% Based on the incremental proximal descent method proposed by Richard et al.~\cite{Richard26}, $\mathbf{Z}_h$ and $\mathbf{Z}_l$ can be updated one more step after Eq.~\eqref{a21} as follows:
\begin{table*}[!t]
	\setlength{\abovecaptionskip}{0cm} 	
	\setlength{\belowcaptionskip}{0.4cm}
	\setlength{\tabcolsep}{1.4mm}
	\caption{The classification results on all eight datasets. The best classification results are bolded, and the second-best results are underlined.} \label{tab-all}
	\centering
 \scriptsize
\begin{tabular}{lcccccccc}
\hline
Dataset & \texttt{Cora} & \texttt{Citeseer} & \texttt{Photo} & \texttt{Computer} & \texttt{Squirrel} & \texttt{Chameleon}& \texttt{Wisconsin} & \texttt{Texas}  \\ \hline
GCN & 85.77 $\pm$ 0.20 & 73.68 $\pm$ 0.31 & 90.54 $\pm$ 0.21  & 82.52 $\pm$ 0.32 & 33.11 $\pm$ 0.46 & 46.53 $\pm$ 0.44& 51.76 $\pm$3.06 & 55.14$\pm$ 5.16 \\
GAT & 86.37 $\pm$ 0.30 & 74.32 $\pm$ 0.27 & 90.09 $\pm$ 0.27   & 81.95 $\pm$ 0.38 & 30.03 $\pm$ 0.25 & 42.93 $\pm$ 0.28& 49.41 $\pm$ 4.09 & 52.16$\pm$ 6.63\\
MLP  & 74.82 $\pm$ 2.22  & 70.94 $\pm$ 0.39   & 78.69 $\pm$ 0.30 &  70.48 $\pm$ 0.28&37.04 $\pm$ 0.46 & 49.67 $\pm$ 0.78& 85.29 $\pm$3.31 & 80.81 $\pm$ 4.75 \\ \hline
GCNII & 88.49 $\pm$ 2.78 & {77.08 $\pm$ 1.21} & 90.98 $\pm$ 0.93 & {86.13 $\pm$ 0.51}& 37.85 $\pm$ 2.76  &  60.61 $\pm$ 2.00 & 80.39 $\pm$ 3.40 & 77.57$\pm$ 3.83\\
APPNP & 87.87 $\pm$ 0.85  & 76.53 $\pm$ 1.33 &91.11 $\pm$ 0.26&81.99 $\pm$ 0.26 &  33.29 $\pm$ 1.72 &  54.30 $\pm$ 0.34  & 70.72 $\pm$ 1.48 & 70.01 $\pm$ 1.59\\
JKNet & {88.93 $\pm$ 1.35} & 74.37 $\pm$ 1.53 &87.70 $\pm$ 0.70 & 77.80 $\pm$ 0.97 & 44.24 $\pm$ 2.11 &  62.31 $\pm$ 2.76  & 82.55 $\pm$ 4.57 & 83.78 $\pm$ 2.21\\ \hline
Geom-GCN & 85.19 $\pm$ 1.13  & {{77.99} $\pm$ {1.23} }& { 91.96 $\pm$ 0.95 } & 83.47$ \pm$ 1.21 &33.32 $\pm$ 1.59  & 60.31 $\pm$ 1.77 & 64.12 $\pm$ 5.02 & 67.57$\pm$ 4.16 \\
H2GCN& 87.81 $\pm$ 1.35 & 76.88 $\pm$ 1.77 & \text { OOM } &  \text { OOM } &37.90 $\pm$ 2.02 & 59.39 $\pm$ 1.98 & 86.67 $\pm$ 4.69 & 84.86 $\pm$ 6.77\\
Ordered GNN & {88.37 $\pm$ 0.75}  & 77.31 $\pm$ 1.73 &93.22 $\pm$ 0.45  &89.03 $\pm$ 1.01 & {62.44$\pm$  1.96} & 72.28 $\pm$  2.29& 88.04 $\pm$ 3.63& 86.22 $\pm$ 4.12\\
GPRGNN & 88.65 $\pm$ 1.37&{{77.99 $\pm$ 1.64}} &91.93 $\pm$ 0.26   & 82.90 $\pm$ 0.37 & {49.93 $\pm$ 1.34} &   {67.48 $\pm$ 1.98}& 83.94 $\pm$ 4.21 & 78.38 $\pm$ 4.36\\
FAGCN& 87.77 $\pm$ 1.69 & 74.66 $\pm$ 2.27 & {91.96 $\pm$ 0.71} &  86.09 $\pm$ 0.40 &40.88 $\pm$ 2.02 &  61.12 $\pm$ 1.95& 75.43 $\pm$ 4.66 &72.12 $\pm$ 4.66  \\
GNN-BC & {88.75 $\pm$ 1.21}  & 76.70 $\pm$ 0.77 &{{93.17} $\pm$ {0.67}}  & {89.60 $\pm$ 0.89} & 61.41 $\pm$  1.55 & {74.63 $\pm$  0.93}& 86.86 $\pm$ 3.88 & 85.01 $\pm$ 3.99\\ 
LHS&88.71 $\pm$ 0.70 & 78.53 $\pm$ 1.50 & \text { NA } & \text { NA }&60.27$\pm$ 1.20 & 72.31 $\pm$ 1.60 & 88.32 $\pm$ 2.30 &86.32 $\pm$ 4.50 \\
PC-Convs& 90.02 $\pm$ 0.62 & \underline{81.76 $\pm$ 0.78} &92.33 $\pm$ 0.65 &  89.90 $\pm$ 0.43 &62.03 $\pm$ 1.55 &  75.03 $\pm$ 1.15& 88.63 $\pm$ 2.75 & \underline{88.11 $\pm$ 2.17} \\
GNN-SATA &\underline{91.24 $\pm$ 0.69}  &{78.33 $\pm$ 0.75}  & \underline{93.62 $\pm$ 0.55 } &\underline{ 91.34  $\pm$ 0.29}  &\underline{65.25 $\pm$ 1.33 } &\underline{ 77.12 $\pm$ 0.96 }& \underline{91.72$\pm$ 3.23} & {87.95 $\pm$ 3.98} \\  \hline
DFGNN &\textbf{92.33 $\pm$ 0.52}  &\textbf{83.00 $\pm$ 0.90}  & \textbf{94.36 $\pm$ 0.61 } &\textbf{ 92.07  $\pm$ 0.33}  &\textbf{70.02 $\pm$ 1.02 } &\textbf{ 81.03 $\pm$ 0.75 } & \textbf{93.98 $\pm$ 2.02} &\textbf{91.02 $\pm$ 3.57}\\ 

\hline
\end{tabular}
\end{table*}

% \begin{table}[t]
% \small
% \centering
% \setlength{\belowcaptionskip}{0.4cm}
% \caption{The statistics of six datasets.}
% \begin{tabular}{cccccc}
% \hline
% Dataset & Nodes & Features &  Classes & Edges & $\mathcal{H}(\mathbf{A})$ \\ \hline
% \multicolumn{1}{c|}{Cora}         & 2708     & 1433  & 7  & 5429 &0.809\\
% \multicolumn{1}{c|}{Citeseer}    & 3327     & 3703  & 6   &4732 & 0.721      \\
% \multicolumn{1}{c|}{Photo}     & 7650     & 745  & 8  &119081&0.824   \\
% \multicolumn{1}{c|}{Computer}     & 13752     & 767  & 10 & 245861 &0.791    \\
% \multicolumn{1}{c|}{Squirrel}      & 5201     & 2089  & 5 & 217073 &0.203  \\ 
% \multicolumn{1}{c|}{Chameleon} & 2277     & 2325  & 5 & 36101 &0.233\\ 
% \hline
% \end{tabular}
% \label{dataset1}
% \end{table}

\section{Experiments}\label{Sec:5} 
This section introduces the datasets, compared methods and experimental parameter configurations for conducting experiments to  evaluate DFGNN. The experimental results are reported and analyzed, and ablation experiments are designed to analyze the importance of the proposed components.
\subsection{Datasets}
We use the eight most commonly used benchmark datasets for comparison. There are four homophilic datasets (\texttt{Cora}, \texttt{Citeseer}, \texttt{Photo}, \texttt{Computer}) and four heterophilic datasets (\texttt{Squirrel}, \texttt{Chameleon}, \texttt{Wisconsin}, \texttt{Texas}). The Homophily Rate (HR) is calculated using the approach in ~\cite{pei2020geom24} and the description of dataset properties is in Table \ref{dataset1}. 
\subsection{Baselines}
Baseline methods include some traditional methods such as GCN~\cite{kipf2017semi3}, GAT~\cite{velivckovic2017graph27}, and MLP. Additionally, methods based on attribute enhancement to alleviate over-smoothing have also been introduced, such as GCNII~\cite{chen2020simple29} APPNP~\cite{gasteiger2018predict32} and JKNet~\cite{xu2018representation33}. Furthermore, some methods based on heterophily have been compared, including structural exploration methods like Geom-GCN~\cite{pei2020geom24}, H2GCN~\cite{zhu2020beyond31}, Ordered GNN~\cite{song2023ordered35}, and LHS~\cite{qiu2024refining30}; frequency-based methods like GPRGNN~\cite{chien2020adaptive15}, FAGCN~\cite{bo2021beyond13}, PC-Conv ~\cite{li2024pc34}; and recently proposed optimization based methods GNNBC~\cite{yang2022graph7} and GNN-SATA~\cite{yang8}.
\subsection{Implementation Details}
The training of the model was conducted on a machine equipped with an NVIDIA RTX 3090 Ti GPU. In line with baseline methods, datasets were randomly split into training, validation, and test sets in proportions of 60\%, 20\%, and 20\%, respectively. And run on more than 10 randomly split test sets like \cite{yang2022graph7}. Accuracy (ACC) served as the evaluation metric for the node classification task. Higher ACC values indicate better performance in correctly classifying nodes. The model used an early stopping strategy~\cite{prechelt2002early36}, with a maximum iteration count of 1000 and a patience of 25. The learning rate of the network was set within the range of $[10^{-5}, 10^{-2}]$, and $K$ was selected from$\{3,4\}$. For a fair comparison, the experimental results of baseline methods are from the original papers. For the datasets without reported results in the baseline methods, we ran the authors' provided code and performed parameter tuning.

\subsection{Main Results}
All experimental results are shown in Table~\ref{tab-all}. It can be seen from Table~\ref{tab-all}   that DFGNN consistently outperforms all other methods across various datasets, achieving an average improvement of 2.21\% over eight datasets. This underscores its robustness and effectiveness in both homophilic and heterophilic datasets. The improvements are particularly significant in heterophilic datasets, where it achieves an average enhancement of 3.46\%. This is a notable achievement, as traditional methods typically struggle in such scenarios. Methods like GCNII and APPNP, which enhance attribute information, show some improvement over traditional methods but still perform poorly on heterophilic datasets. These methods overly rely on given graph relationships, neglecting the inaccuracies and interference between topology and attributes. DFGNN mitigates this interference by establishing correlations between corresponding frequency bands of topology and attributes.

We also analyzed several competitive state-of-the-art heterophilic graph methods. PC-Conv tightly couples the low-frequency and high-frequency filtering processes for homophilic and heterophilic graphs. However, it does not fine-tune for each frequency band, limiting the model's ability to handle different graph data types. In contrast, DFGNN can independently optimize different frequency bands simultaneously, enhancing the overall flexibility of the model. Compared to LHS, which uses complex self-expression and contrastive learning mechanisms, and GNN-SATA, which explores latent structures by learning two mask matrices, DFGNN explicitly filters and combines frequency components. This provides a more direct approach to handling graph heterophily and homophily, achieving better results with enhanced interpretability.

Additionally, methods like PC-Conv, LHS, and GNN-SATA primarily improve the robustness of their models to graph structure. In contrast, DFGNN explicitly removes noise and redundant information across different frequency bands through sparse and low-rank constraints. This approach enhances robustness to both structural and attribute noise. Most methods require hyperparameters in their objective functions to be personalized for different datasets. However, DFGNN employs an adaptive adjustment strategy, allowing the network to be self-aware based on downstream tasks. Consequently, DFGNN achieves significant advantages over the current best methods.

\subsection{Ablation Study}
To validate the effectiveness of each component in the DFGNN model, we designed three ablation experiments by removing the low-pass filter, high-pass filter, and sparse and low-rank constraints, respectively, as follows:

\textbf{DFGNN-H} denotes the removal of the high-pass filter to evaluate the contribution of high-frequency components to the model. The optimization objective function is:
           \begin{equation}
      	\begin{split}
&\min _{\mathbf F, \mathbf{Z}_l}\left\| \mathbf{F}-\mathbf{X}\right\|_F^2+ \alpha_1 (\operatorname{tr}\left(\mathbf{Z}_l^{\top} {\mathbf{L}_{l}}\mathbf{Z}_l\right)+\left\| \mathbf{F}_{l}-\mathbf{Z}_l \right\|_F^2)
\\& +\alpha_3\|\mathbf{Z}_l\|_*
           	\end{split}
            \label{ablation1}
      \end{equation}

\textbf{DFGNN-L} represents the removal of the low-pass filter to evaluate the contribution of the low-frequency component to the model's performance. The optimization objective function is:
                 \begin{equation}
      	\begin{split}
&\min _{\mathbf F, \mathbf{Z}_h}\left\| \mathbf{F}-\mathbf{X}\right\|_F^2+ \alpha_2 (\operatorname{tr}\left(\mathbf{Z}_h^{\top} {\mathbf{L}_{h}}\mathbf{Z}_h\right)+\left\| \mathbf{F}_{h}-\mathbf{Z}_h \right\|_F^2)
\\&+\alpha_4\|\mathbf{Z}_h\|_1.
           	\end{split}
            \label{ablation2}
      \end{equation}
\begin{table}[!t]
	\caption{The experimental results of three ablation modules and DFGNN. The best classification results are bolded.} \label{tababl}
	\centering
 \scriptsize
\begin{tabular}{c|ccc|c}
\hline
Dataset & DFGNN-H & DFGNN-L & DFGNN-C & {DFGNN} \\ \hline
\texttt{Cora} & 91.05 & 85.40 & 92.10 & \textbf{92.33} \\ \hline
\texttt{Citeseer} & 81.05 & 60.30 & 82.20 & \textbf{83.00} \\ \hline
\texttt{Photo} & 92.58 & 90.22 & 93.75 & \textbf{94.36} \\ \hline
\texttt{Computer} & 89.63 & 87.03 & 90.13 & \textbf{92.07} \\ \hline
\texttt{Chameleon} & 78.21 & 78.64 & 78.13 & \textbf{81.03} \\ \hline
\texttt{Squirrel} & 66.09 & 68.03 & 68.04 & \textbf{70.02} \\ \hline
\texttt{Wisconsin} & 86.00 & 88.15& 92.82 & \textbf{93.98} \\ \hline
\texttt{Texas} & 83.78& 86.42 & 86.48 & \textbf{91.02} \\ \hline
\end{tabular}
\end{table}
\textbf{DFGNN-C} represents the model without the sparse and low-rank constraints to evaluate the contribution of the constraints to the model's performance. The optimization objective function is:
           \begin{equation}
      	\begin{split}
&\min _{\mathbf F, \mathbf{Z}_l,\mathbf{Z}_h}\left\| \mathbf{F}-\mathbf{X}\right\|_F^2+ \alpha_1 (\operatorname{tr}\left(\mathbf{Z}_l^{\top} {\mathbf{L}_{l}}\mathbf{Z}_l\right)+\left\| \mathbf{F}_{l}-\mathbf{Z}_l \right\|_F^2)
\\&+ \alpha_2 (\operatorname{tr}\left(\mathbf{Z}_h^{\top} {\mathbf{L}_{h}}\mathbf{Z}_h\right)+\left\| \mathbf{F}_{h}-\mathbf{Z}_h \right\|_F^2)
           	\end{split}
            \label{ablation3}
      \end{equation}

As seen in Table~\ref{tababl}, the results of all three variants are lower than DFGNN,  indicating the importance of each module. For the homophilic datasets, DFGNN-H performs relatively well, indicating that low-frequency information plays a crucial role in homophilic graphs. Even after removing the high-frequency part, the model can still capture the homophilic nature of nodes. For the heterophilic datasets, DFGNN-H shows a significant performance drop. Heterophilic graphs require high-frequency information to distinguish between neighbours of different classes, and removing the high-frequency part limits the model's ability to capture node heterogeneity.

DFGNN-L performs poorly on homophilic graphs, especially on the Citeseer dataset, showing a significant decline in performance. This indicates that high-frequency information contributes little to homophilic graphs and may even introduce noise. In contrast, DFGNN-L performs relatively well on heterophilic graphs because high-frequency information is more critical for distinguishing different classes of nodes. Although there is a performance drop, it is smaller than that observed in homophilic graphs.

DFGNN-C shows performance drops across all datasets, indicating that the sparse and low-rank constraint are crucial for improving model performance by removing noise and redundant information, thereby extracting effective dual-frequency representations.
\begin{figure*}[!t]
\centering

     \subfigure[Cora-SATA\newline sc=0.131]{\label{2}
 \includegraphics[width=0.15\textwidth]{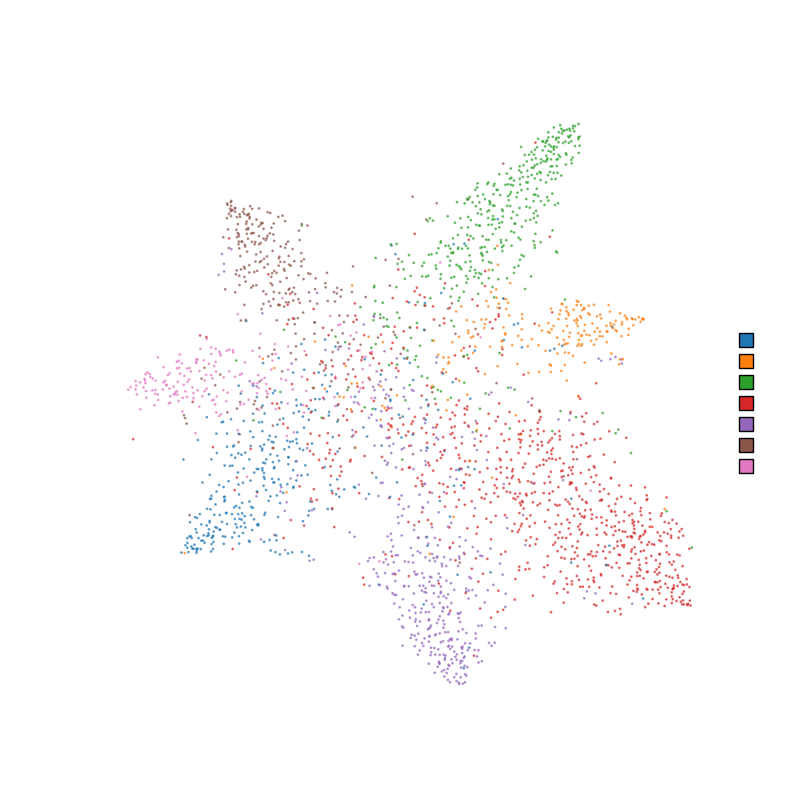}}
 \subfigure[Citeseer-SATA \newline sc=0.256]{\label{4}
 \includegraphics[width=0.15\textwidth]{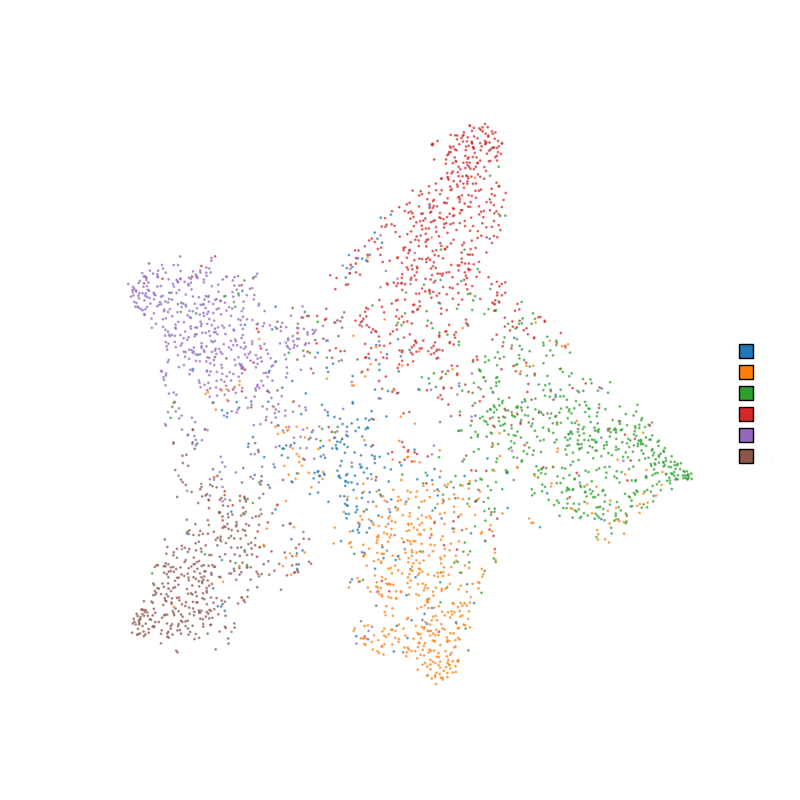}} 
  \subfigure[Photo-SATA\newline sc=0.303]{\label{6}
 \includegraphics[width=0.15\textwidth]{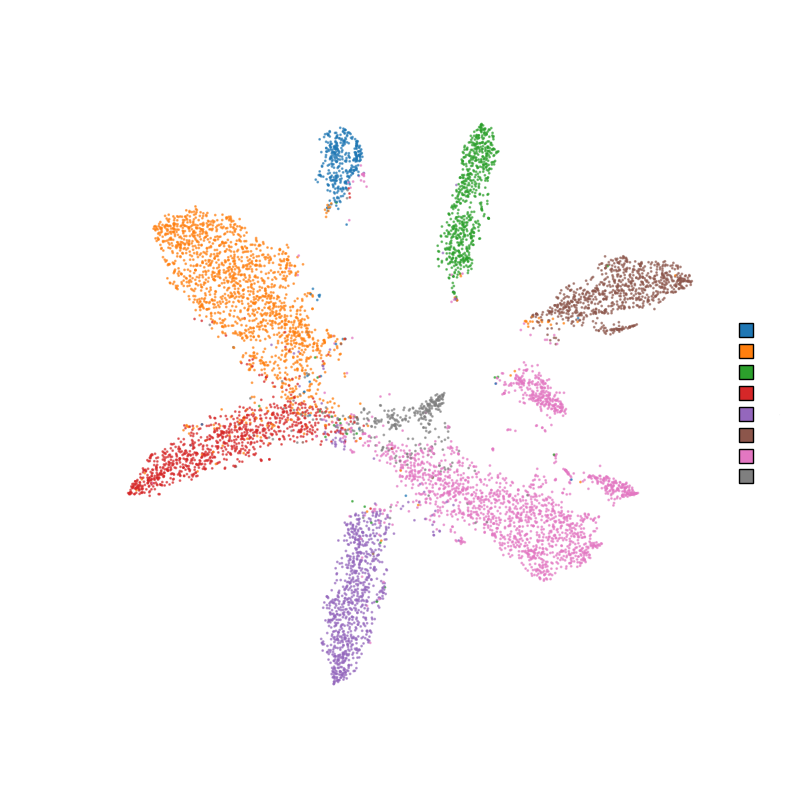}}
   \subfigure[Computer-SATA\newline sc=0.139]{\label{8}
 \includegraphics[width=0.16\textwidth]{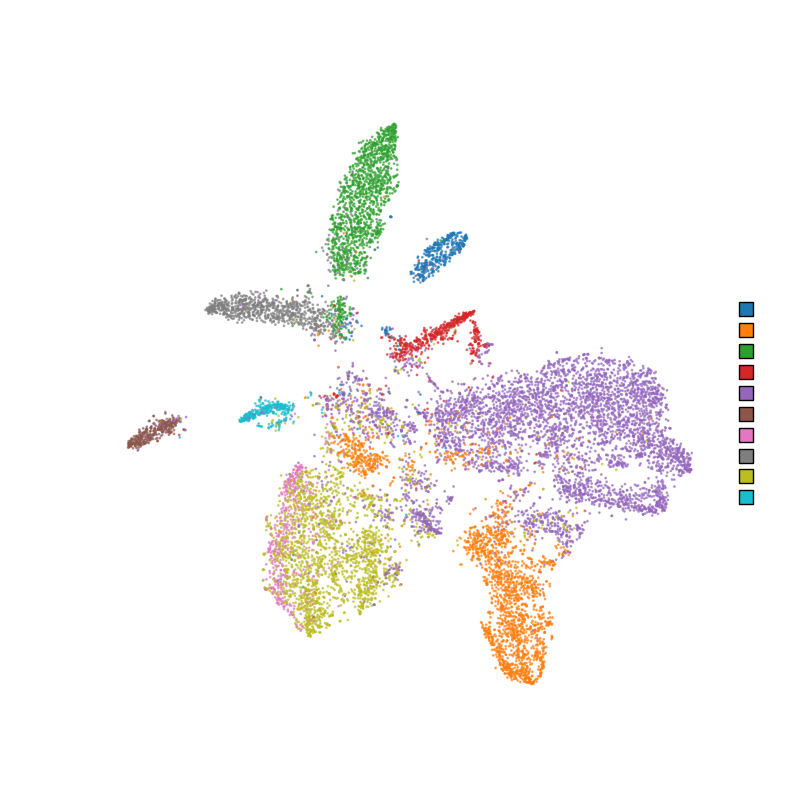}}
  \subfigure[Squirrel-SATA\newline sc=0.153]{\label{10}
 \includegraphics[width=0.15\textwidth]{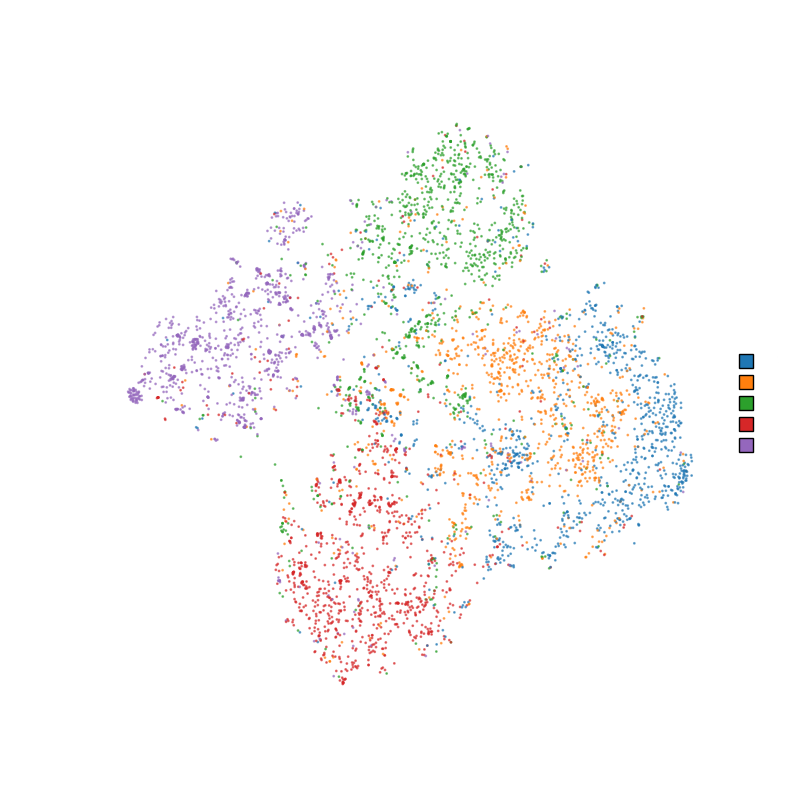}} 
  \subfigure[Chameleon-SATA\newline sc=0.216]{\label{12}
 \includegraphics[width=0.17\textwidth]{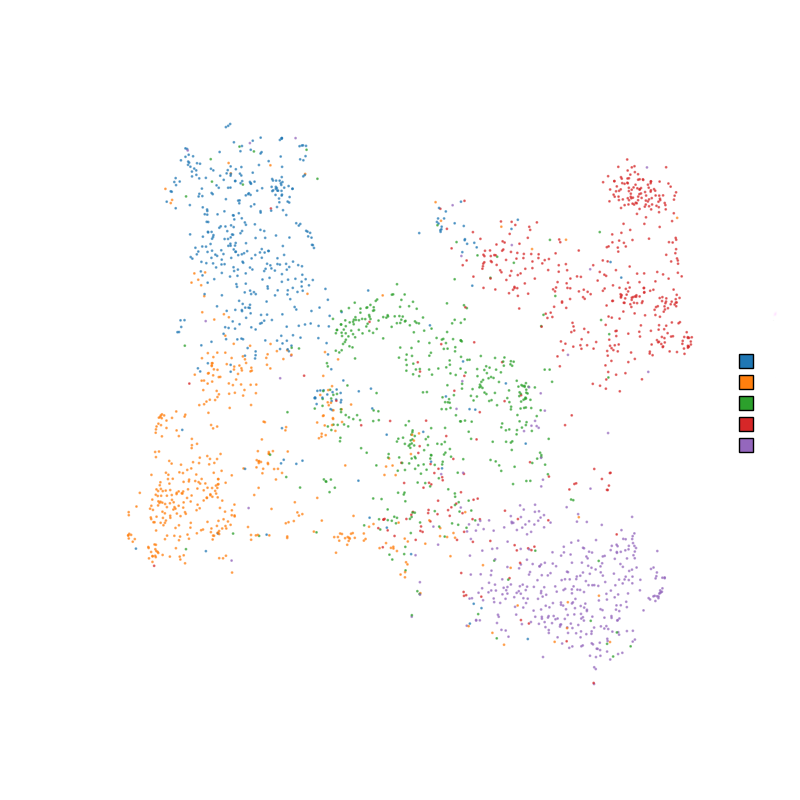}}
%\quad\subfigure[]{\label{acc_para5}\includegraphics[width=1.4in]{coil100_para}}
 \subfigure[Cora-DFGNN\newline sc=0.241]{\label{1}
 \includegraphics[width=0.15\textwidth]{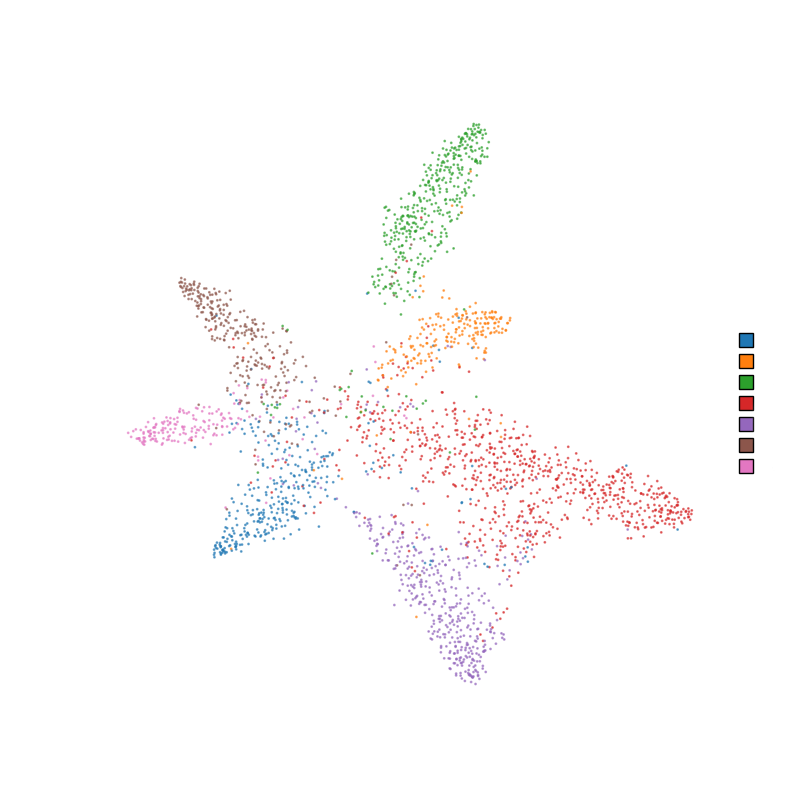}} 
   \subfigure[Citeseer-DFGNN\newline sc=0.262]{\label{3}
 \includegraphics[width=0.15\textwidth]{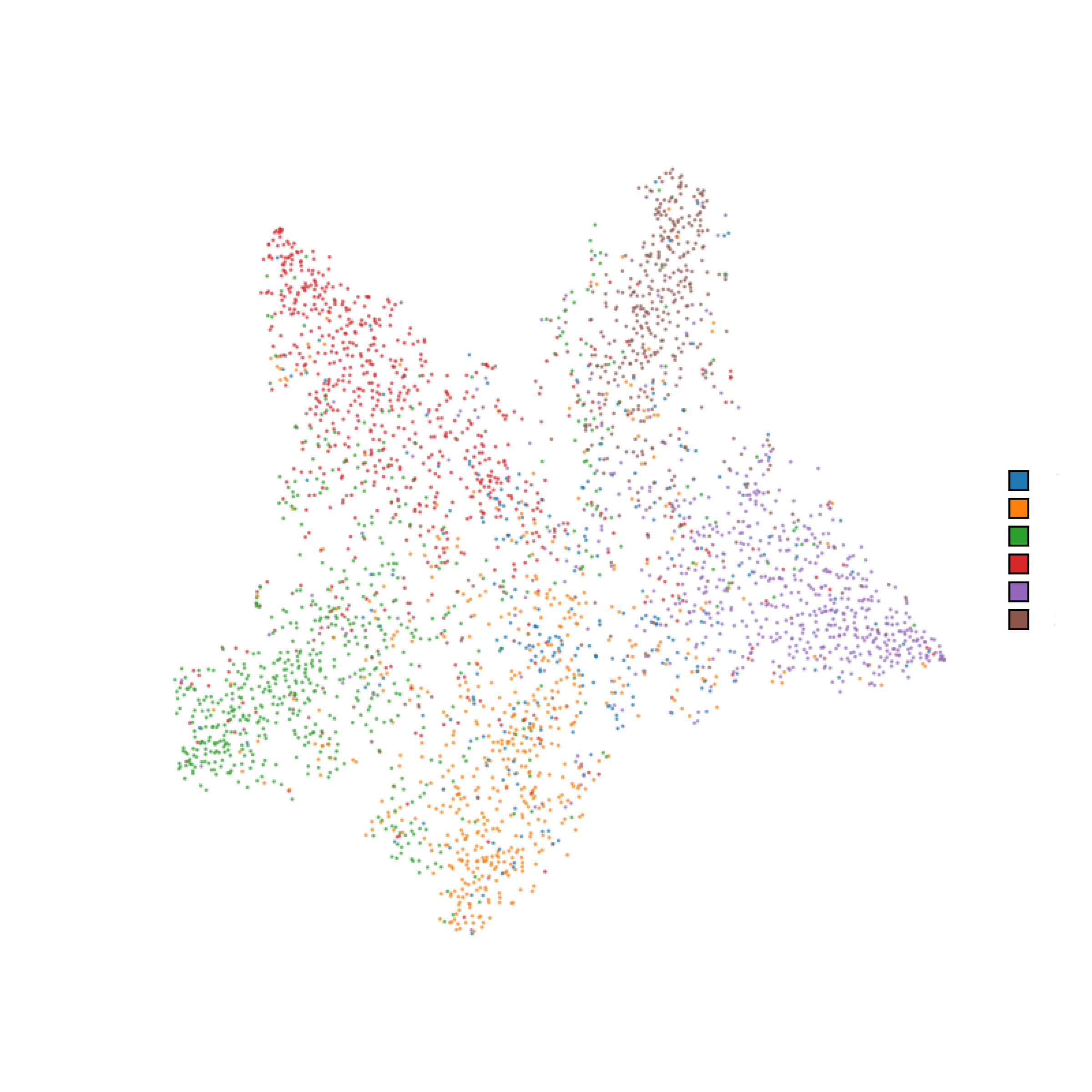}}
  \subfigure[Photo-DFGNN\newline sc=0.320]{\label{5}
 \includegraphics[width=0.15\textwidth]{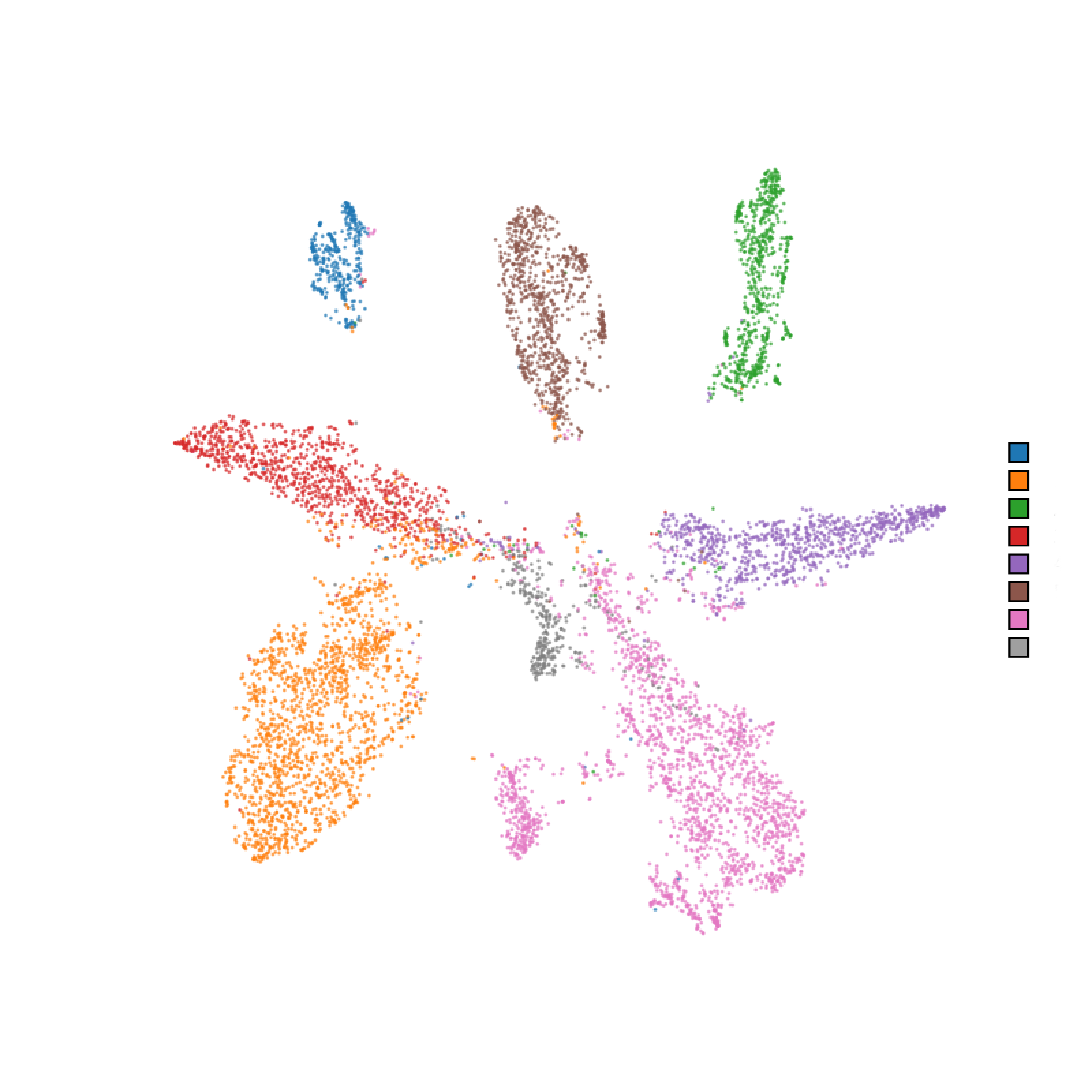}}
   \subfigure[Computer-DFGNN\newline sc=0.147]{\label{7}
 \includegraphics[width=0.16\textwidth]{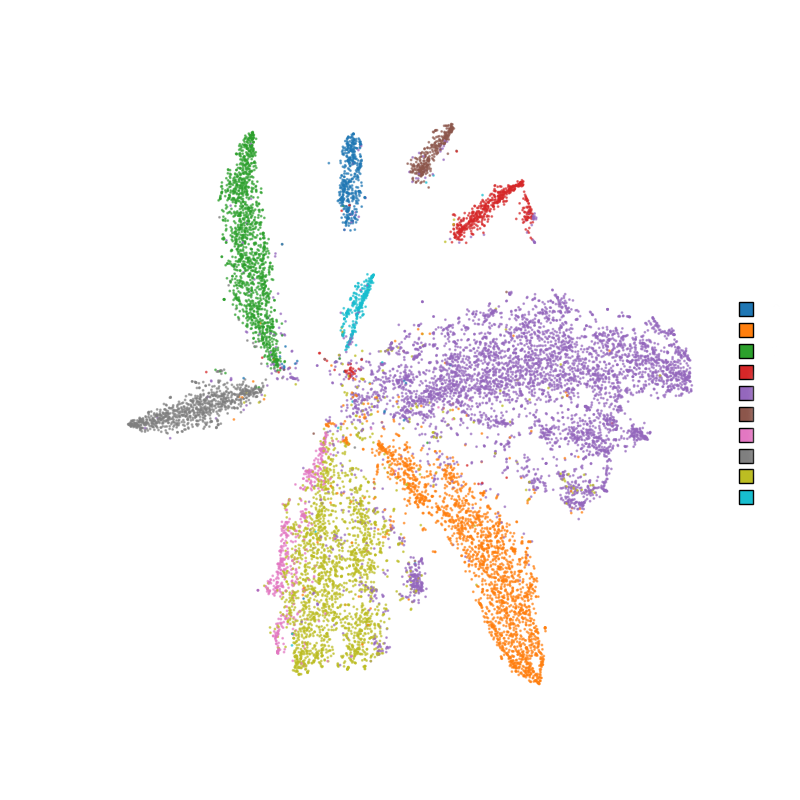}} 
    \subfigure[Squirrel-DFGNN\newline sc=0.177]{\label{9}
 \includegraphics[width=0.15\textwidth]{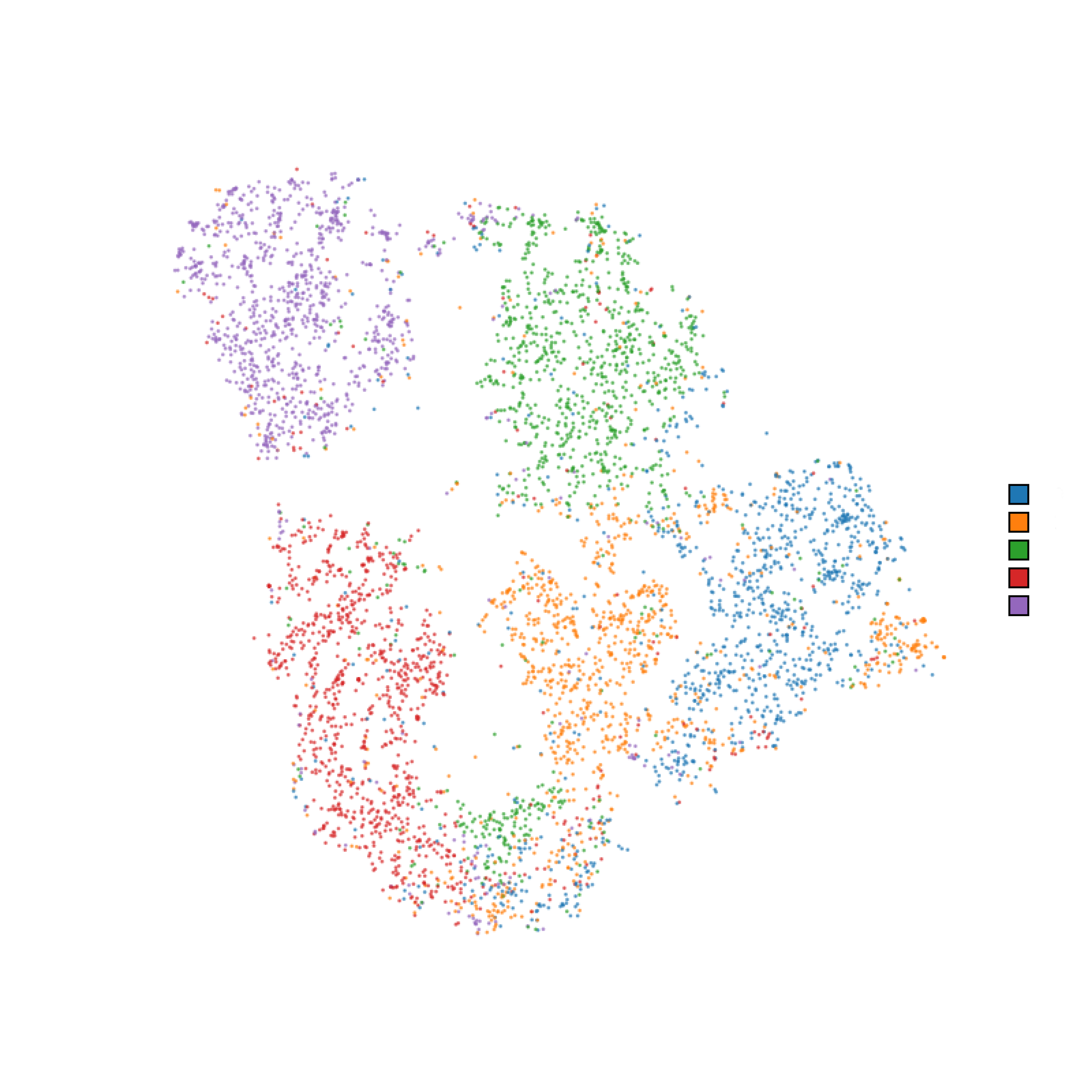}}
\subfigure[Chameleon-DFGNN\newline sc=0.234]{\label{11}
 \includegraphics[width=0.17\textwidth]{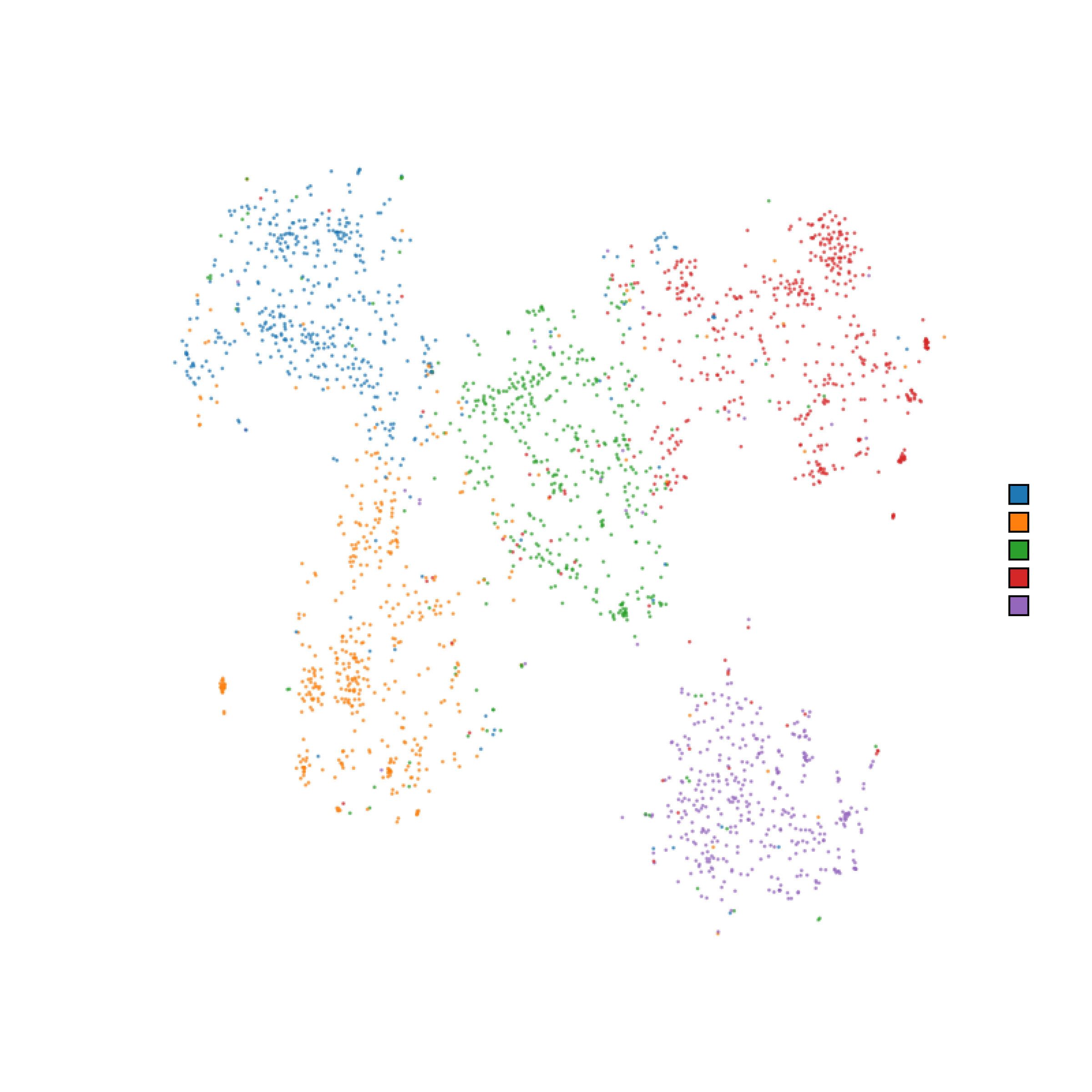}}
\caption{The visualization of classification results for DFGNN and GNN-SATA approaches}. \label{Fig_vis1}
\end{figure*}
\begin{figure}[!t]
\centering

     \subfigure[Cora]{\label{o1}
 \includegraphics[width=0.18\textwidth]{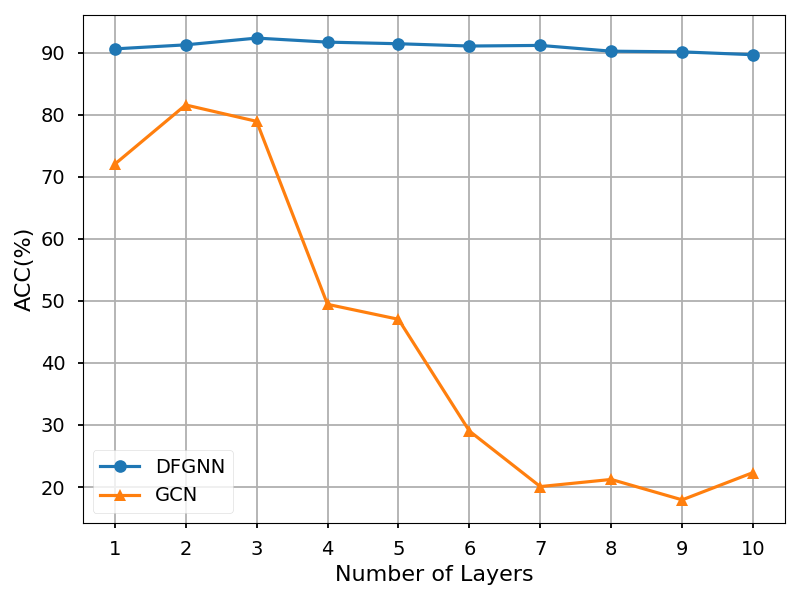}}
 \subfigure[Citeseer]{\label{o2}
 \includegraphics[width=0.18\textwidth]{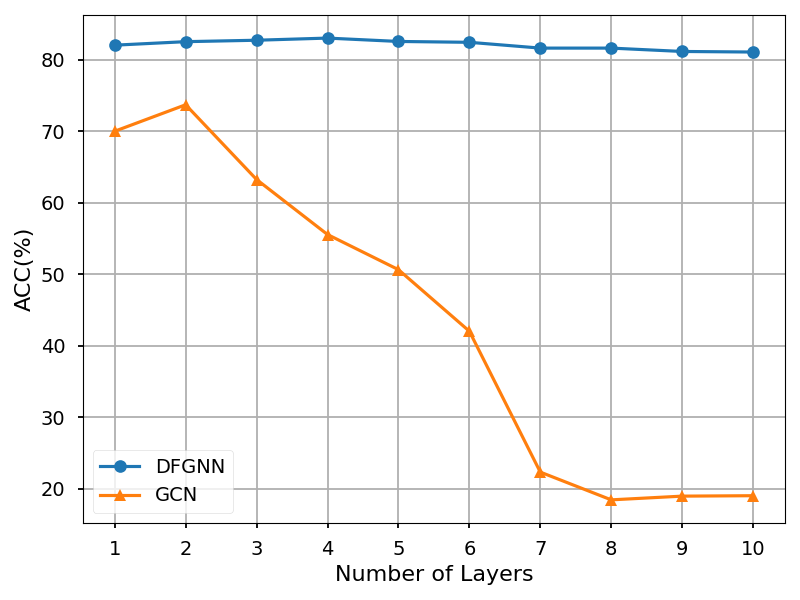}} 
 %  \subfigure[Photo]{\label{o3}
 % \includegraphics[width=0.2\textwidth]{photoover.jpg}}
 %   \subfigure[Computer]{\label{04}
 % \includegraphics[width=0.2\textwidth]{computerover.jpg}}
  \subfigure[Squirrel]{\label{o5}
 \includegraphics[width=0.18\textwidth]{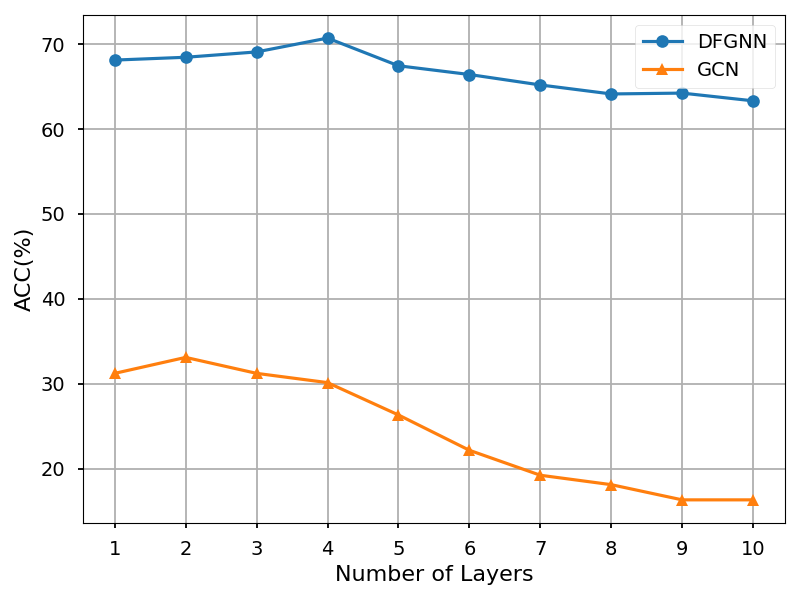}} 
  \subfigure[Chameleon]{\label{o6}
 \includegraphics[width=0.18\textwidth]{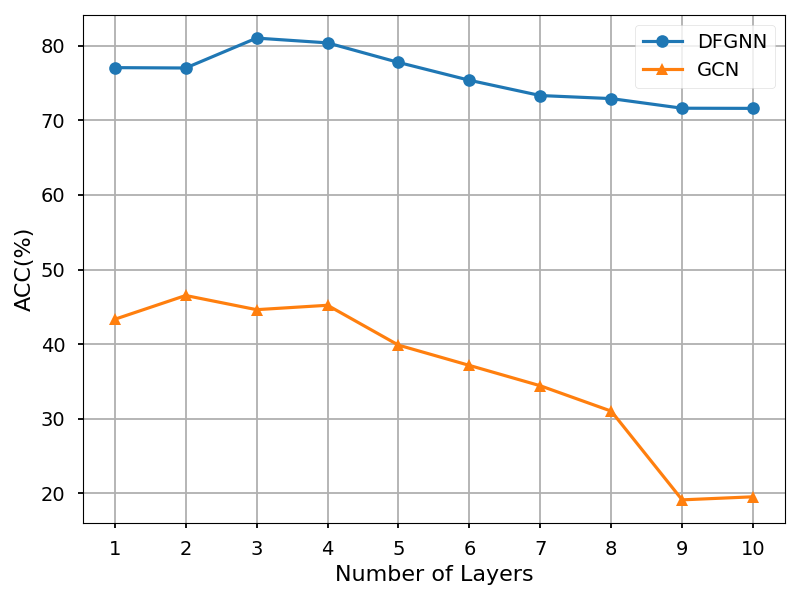}}
\caption{ Classification results comparison of GCN and DFGNN  with different number of layers.} \label{Fig_oversoooth}
\end{figure}

\subsection{Alleviating Over-smoothing Problem}
To verify DFGNN's effectiveness in resisting over-smoothing, we tested the performance of DFGNN and GCN on four datasets at various model depths. The results, shown in Figure.~\ref{Fig_oversoooth}, indicate that GCN's performance declines rapidly as the number of layers increases, highlighting severe over-smoothing issues. In contrast, DFGNN maintains stable performance across different depths and significantly outperforms GCN. The reasons for this can be summarized as follows:

First, DFGNN employs a dual-frequency filtering design with low-pass and high-pass filters. The low-pass filter captures smooth global features, while the high-pass filter retains local detail features. This design ensures that the model preserves both global information and local distinctions when processing homophilic and heterophilic graphs, reducing the risk of over-smoothing.

Additionally, DFGNN establishes new relationships between topology and attributes to mitigate their interference. This approach captures node attribute representations while retaining graph structure information, thus reducing the impact of over-smoothing.

Finally, the frequency-specific constraints introduced in DFGNN further reduce noise and redundant information, thereby  helping to maintain the discriminative power of node representations in deeper networks.

\subsection{Results Visualization}

To validate the effectiveness of the proposed DFGNN model, we used T-SNE~\cite{van2008visualizing37} to visualize the node classification results of both DFGNN and the latest GNN-SATA (Figure.~\ref{Fig_vis1}) on six datasets. The results for GNN-SATA are labelled (-SATA), while those for DFGNN are labelled (-DFGNN). We also employed the silhouette coefficient (sc)~\cite{rousseeuw1987silhouettes38} to accurately measure the differences between the two models.
DFGNN consistently exhibits higher silhouette coefficients across all datasets, indicating superior performance. This improvement is attributed to the sparse low-rank constraints on different frequency domain representations, which filter out noise and redundant information. This enhancement makes DFGNN more robust and leads to more accurate and compact results compared to GNN-SATA.
\section{Conclusion}\label{Sec:7}

This paper introduces DFGNN, a novel graph neural network model designed for semi-supervised node classification. From an optimization perspective, DFGNN incorporates low-pass and high-pass filters at the topological level to capture both smooth and detailed features. The model dynamically adjusts filtering rates to handle graphs with different properties, while applying frequency-specific constraints to reduce noise and redundancy. Ablation experiments demonstrate the necessity of combining dual-frequency information and applying frequency-specific constraints. Additionally, DFGNN  establishes dynamic correspondences between topological and attribute frequency bands to mitigate representation distortion caused by their interference. Experimental results demonstrate DFGNN's superior performance on homophilic and heterophilic graphs.

%Bibliography
\bibliographystyle{unsrt}

\end{document}